# Çok Kipli Duygu Tanıma İçin Zaman Bazlı Modellerin Karşılaştırılması

# A Comparison of Time-based Models for Multimodal Emotion Recognition

Ege Kesim, Selahattin Serdar Helli, Sena Nur Cavsak
ege.kesim1@huawei.com, serdar.helli1@huawei.com, sena.nur.cavsak@huawei.com
Huawei Turkey Research and Development Center, Istanbul

*Özetçe*

**Duygu tanıma, insan-bilgisayar etkileşim alanında önemli bir araştırma konusu haline gelmiştir. Duyguları anlamak için ses ve videolar üzerine yapılan çalışmalar, temel olarak yüz ifadesini analiz etmeye odaklanmış ve 6 temel duyguyu sınıflandırmıştır. Bu çalışmada, çok kipli duygu tanımada farklı sekans modellerinin performansları karşılaştırılmıştır. Ses ve görüntüler önce çok katmanlı CNN modelleri tarafından işlenmiş ve bu modellerin çıktıları çeşitli sekans modellerine beslenmiştir. Sekans modeli olarak GRU, Transformer, LSTM ve Maksimum Havuzlama kullanılmıştır. Bütün modellerin kesinlik, duyarlılık ve F1 Skor değerleri hesaplanmıştır. Deneylerde çok kipli olan CREMA-D veri seti kullanılmıştır. CREMA-D veri kümesinin karşılaştırma sonucunda, F1 skorda en iyi sonucu gösteren 0.640 ile GRU bazlı mimari, kesinlik metriğinde 0.699 ile LSTM bazlı mimari, duyarlılıkta ise zaman içinde maksimum havuzlama bazlı mimari 0.620 ile en iyi sonucu göstermiştir. Sonuç olarak, karşılaştırılan sekans modellerinin birbirine yakın performans verdiği gözlenmiştir.**

*Anahtar Kelimeler — Otomatik ses-video duygu tanıma (AVER), Duygu tanıma, Mel-frekans kepstrum katsayıları, Çok kipli füzyon, Transformers, Derin öğrenme.*

*Abstract*

**Emotion recognition has become an important research topic in the field of human-computer interaction. Studies on sound and videos to understand emotions focused mainly on analyzing facial expressions and classified 6 basic emotions. In this study, the performance of different sequence models in multi-modal emotion recognition was compared. The sound and images were first processed by multi-layered CNN models, and the outputs of these models were fed into various sequence models. The sequence model is GRU, Transformer, LSTM and Max Pooling. Accuracy, precision, and F1 Score values of all models were calculated. The multi-modal CREMA-D dataset was used in the experiments. As a result of the comparison of the CREMA-D dataset, GRU-based architecture with 0.640 showed the best result in F1 score, LSTM-based architecture with 0.699 in precision metric, while sensitivity showed the best results over time with Max Pooling-based architecture with 0.620. As a result, it has been observed that the sequence models compare performances close to each other.**

## I. GİRİŞ

Goleman duyguyu tanımlarken, insanı harekete geçmeye sevk eden ve yaşamın güçlükleri ile baş etmesini sağlayan hisler olarak tarif eder ve hayatın her alanında büyük etkiye sahip olduğunu belirtir. İnsanların duygusal durumlarını doğru bir şekilde tanıması, insan ilişkilerinde hayati bir rol taşımaktadır. İnsan ilişkilerinde bir kişinin duygusal durumunu karşı tarafa doğru bir şekilde aktarabilmesi, iletilen mesajın daha iyi anlaşılmasına neden olur. Günümüzde duygu tanıma, çeşitli amaçlar için kullanılmaktadır. Okulda [17, 18], akıllı kartlarda, eğlence ve sağlık hizmetlerinde [11], robot teknolojisinde ve güvenlik kontrolünde [19] duygu tanıma uygulanmaktadır. Kurumsal uygulamalarda ise, perakende, medya, insan kaynakları, müşteri iletişimini içeren çağrı merkezlerinde, kişisel asistanlarda ve akademide duygu analitiğinin çok sayıda kullanım senaryosu vardır.

Yapay zeka teknolojileri ile duygu tanıma, insan duygularını algılama ve anlama sürecinin simülasyonuna odaklanmaktadır. Şu anda, duygu tanıma hala zorlu bir konudur ve araştırmacıların odağındadır. Sağlam otomatik duygu tanıma yöntemleri geliştirmek için artan bir talep vardır. Bu nedenle ses, video, konuşma veya yüz ifadelerinden insan duygularını tanımaya çalışan araştırmaların sayısında son yıllarda önemli bir artış görülmüştür.

Sesler duygusal ifade için önemli bir yöntemdir. Konuşma, duygularla zenginleştirilmiş bir iletişim kanalıdır. Duygu tanımada öznitelik çıkarımı sıklıkla kullanılmaktadır. Bu kullanılan öznitelikler arasında MelFrequency Cepstral katsayıları (MFCC) [1,2] ve Mel-Spectrogram [3,4] gibi klasik öznitelik çıkarımları yada derin öğrenme tabanlı Wav2vec 2.0 [5] yöntemleri kullanılmaktadır [6,7].

Duygu tanımada görüntüler de kullanılabilmektedir. Bu görüntüler genellikle kişilerin yüz görüntülerinden oluşmaktadır. Genellikle görüntülerden duygu tanıma yöntemi yüz algılama ve sınıflandırma olarak iki aşamadan oluşmaktadır.

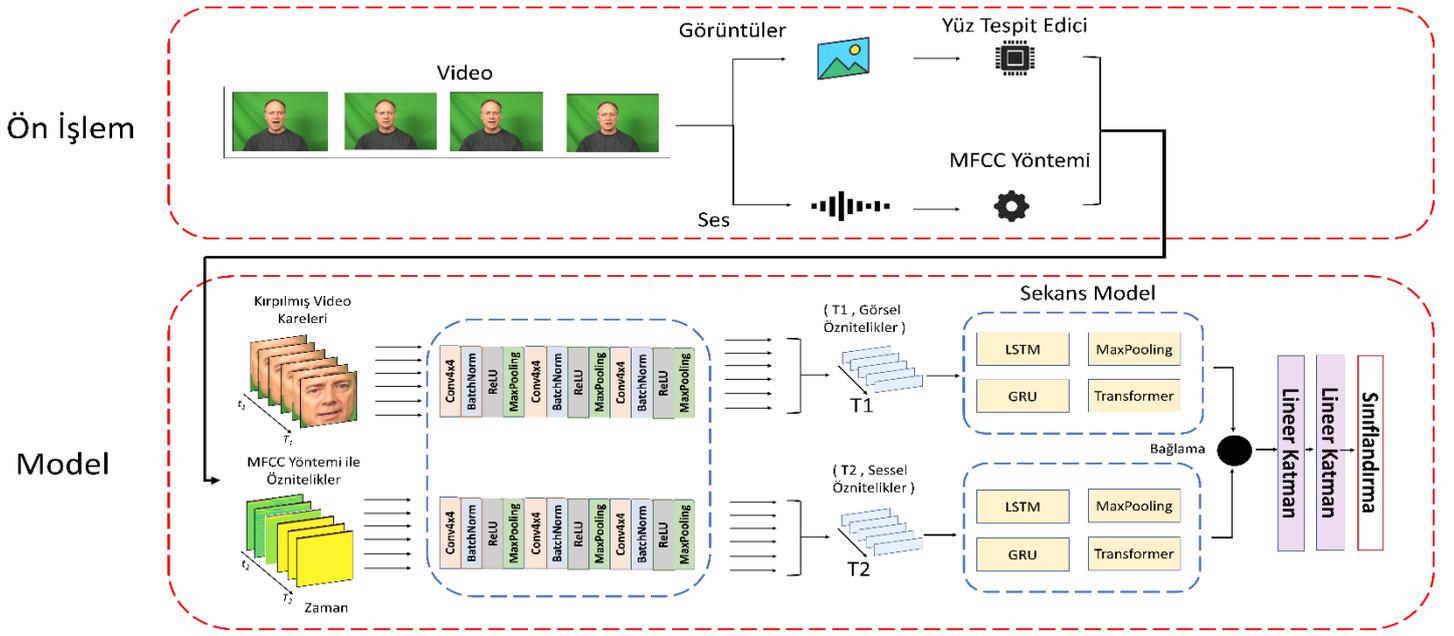

Şekil 1. Sunulan Modelin Diagramı

Görüntü işleme alanındaki başarılarından dolayı CNN [8, 9, 10] modelleri duygu tespitinde sıklıkla kullanılmaktadır.

Çok kipli duygu tanıma, çoklu modalitelerden (ses, video, metin, fizyolojik ve diğerleri) yararlanarak duygu tahmini yapılmasıdır. [11]'deki çalışmalarında, yazarlar çok kipli duygu tanıma konusu üzerinde çalışmışlardır. Bu çalışmada kullanılan modelin görsel ve işitsel verilerden çıkardığı bilgiler arasındaki uzaklık optimize edilerek iki kipten benzer bilgiler çıkarılması sağlanmıştır. Ayrıca, üçlü kayıp fonksiyonu (Triplet Loss) kullanarak model performansı arttırılmıştır. CREMA-D [12] veri kümesinde %74'lük bir doğruluk elde ettiklerini raporlamışlardır.

[13]' da yazarlar, kısa video kliplerini kapsayan ve çiftli ses-görsel zaman pencerelerinden oluşan yeni bir çerçeve önermektedirler. Ses ve görüntü için çıkarılan gömüler (embbeding) iki ayrı transformer'ın [15] kodlayıcıları tarafından işlenmektedir. Bu yeni çift taraflı çerçeve, duygu tanıma için ses-görsel ipuçlarının zamansal sunumunu incelemeyi amaçlamaktadır. CREMA-D veri kümesinde %67.2'lik bir doğruluk elde ettiklerini raporlamışlardır.

[14]'da yazarlar AuxFormer adlı Transformer bazlı yeni bir model yapısı sunmuşlardır. Modelde hem çok kip için hem de her kip için ayrı kayıp fonksiyonları kullanılmaktadır. Tek kipli kayıp fonksiyonlarının eklenmesi ana ağdaki belgelerin kaybolmasını engellemektedir. Yazarlar, CREMA-D veri kümesinde yaptıkları test sonuçlarında makro ve mikro F1 Skoru için sırasıyla %70.4'lük ve % 76.5 performans elde etmişlerdir.

Bu çalışmada, derin öğrenme yöntemi ile ses ve görüntüler üzerinden çok kipli duygu tanıma yapılmıştır. Ses ve görüntüler önce çok katmanlı evrişimsel sinir ağ modelleri (CNN) tarafından işlenmiş ve bu modellerin çıktıları GRU, Transformer, LSTM ve zaman boyunca maksimum havuzlama modellerine beslenmiştir. Bu dört farklı sekans modeli karşılaştırılmıştır.

Makalenin devamında, Bölüm II'de kullanılan modeller anlatılmakta, Bölüm III'de veri kümesi ve yapılan deneyler bulunmakta, Bölüm IV'te ise sonuç kısmı ile çalışma tamamlanmaktadır.

## II. YÖNTEM

### A. Uzun-Kısa Vadeli Bellek (LSTM)

LSTM, bir tekrarlayan sinir ağı (Recurrent Neural Network- RNN) mimarisidir ve RNN'in karşılaştığı sorunları çözmek için geliştirilmiş bir yapay sinir ağıdır. LSTM ağları gradyan kaybolması sorununun üstesinden gelebilmek için bilgi akışını düzenleyen ve kontrol eden kapılardan oluşturmaktadır.

### B. Geçitli Tekrarlayan Ünite (GRU)

GRU, tekrarlayan sinir ağlarında sık görülen gradyan sorununu çözmek amacıyla önerilmiştir. GRU, LSTM'e göre daha az parametreye sahip olduğundan daha küçük ve seyrek veri kümelerinde daha iyi performans göstermektedir.

### C. Transformer

Transformer, Doğal Dil İşleme (NLP) mimarilerinin temel yapı taşını oluşturmaktadır ve dikkat mekanizmasına (attention-based) dayanmaktadır. Transformer, birbirine bağlı ve birbirlerini etkileyen uzak veri öğeleri dizisinde ince bağlantıları algılamak için kullanılmaktadır. Bu model paralel olarak çalıştırmaya daha uygun ve uzun metinlerdeki anlamı daha kolay hatırlayabilmektedir.

### D. Maksimum Havuzlama (Max Pooling)

Maksimum havuzlama, belli bir alandaki en büyük değeri alan bir filtre gibi çalışmaktadır. Veri boyutu azaltılarak hem gereken işlem gücü hem de yakalanan gereksiz özellikler azaltılır ve böylece daha önemli özelliklere odaklanılır.

## E. Önerilen Model

Bu çalışmada, aynı ana omurgaya sahip dört farklı mimariyi karşılaştırıyoruz. Ana omurga, ses ve video girdileri için 3 katmanlı iki CNN'den oluşur. Her katman sırasıyla bir konvolüsyon, toptan normalleştirme (Batch Normalization), doğrultulmuş doğrusal ünite (RELU) ve maksimum havuzlama katmanlarından oluşur. Her konvolüsyon katmanın filtre sayısı 32 olarak sabitlenmiştir. Bu CNN'lerin çıktı vektörleri daha sonra ayrı ayrı bir lineer katmana ve ardından her bir çıktı vektörü bir sekans model ile beslenir. Sekans modeli GRU, Transformer, LSTM veya zaman boyunca maksimum havuzlama olarak 4 farklı yapıda incelenmiştir. Sekans modellerinde, her bir modaliteden gelen zaman çıktısı alınıp birleştirilir. LSTM ve GRU da son zaman adım çıktısı alınır. Transformer modelinde ise global maksimum havuzlama katmanı son çıktı ile beslenir. Zaman boyunca maksimum havuzlama ağı için tek fark, son zaman adımını kullanmak yerine, CNN'lerin çıktılarına zaman boyunca maksimum havuzlama işlemi uygulamaktadır. Bu işlem, her bir CNN çıktısına ayrı ayrı uygulanır. Ardından çıktılar tekrar birleştirilir ve bir lineer katmana beslenir. Daha sonra elde edilen birleştirilmiş vektör, üç lineer katmanla beslendikten sonra, sınıf tahminini veren son Softmax katmanından son sonuç alınır.

Transformer içindeki beslemeli ileri besleme ağının (feed forward layer) gizli katman boyutu ve her belirtecin gömme (embedding) boyutu 64'tür. Ayrıca, transformer bloğu, öz-dikkat mekanizması (self-attention) 4 dikkat başlığına sahiptir. LSTM ve GRU'da ise birim gizli katman boyutu 128 olarak sabitlenmiştir. Tam bağlı blokta (the fully connected), CNN çıktısına global maksimum havuzlama uygulanır, ardından üç tam bağlı katman uygulanır. Katmanların birim boyutları sırasıyla 128, 64 ve sınıf sayısıdır.

## III. DENEYLER

### 1. Veri Kümesi

Bu çalışmada CREMA-D veri kümesini kullanılmıştır. Veri kümesi, 6 temel duyguyu (mutlu, kızgın, üzgün, nötr, iğrenme ve korku) gösteren profesyonel oyuncuların 7442 klibinden oluşur. Her klipte bir oyuncu, verilen bir cümle söylerken duygu kategorilerinden birini sergiler. Kliplerin süresi bir ila beş saniye arasında değişir. Videolardaki farklı cümle sayısı on iki ve toplam oyuncu sayısı 91'dir. Duygu tanımadaki insan performansı veri kümesi üzerinde sadece ses, sadece görüntü, hem ses hem de görüntü kullanılarak ölçülmüştür. Bu durumlar için insan performansı sırasıyla %40,9, %58,9 ve %63,6 olarak ölçülmüştür.

### 2. Veri Ön İşleme

#### Video

Bu çalışmada, görsel özellikler olarak oyuncuların görüntülerdeki yüz görüntüleri içerecek şekilde görüntüler kırpılmış ve 64x64 boyuta indirilerek kullanılmıştır. Yüzlerin tespit edilmesinde Dlib [16] kütüphanesi kullanılmıştır. Ayrıca zaman adımlarının sayısı 50'ye sabitlenmiştir.

#### Ses

Bu çalışmada, sesi temsil etmek için Mel Frekans Cepstral Katsayıları (MFCC) kullanıldı. 22 kHz örnekleme oranında girdi sinyallerini kullanarak çerçeve uzunluğu 2048 ve atlama uzunluğu 512 kullanarak ses kaynaklarından MFCC çıkarıldı. Bu işlemlerde Librosa kütüphanesi kullanıldı. Tüm kliplerin uzunluğu 5 saniye olacak şekilde dolduruldu. Elde edilen öznitelikler 216x40 boyutlu bir matris şeklindedir. Daha sonra, bu öznitelik 40 boyutunda bir kayan pencere ve 20 adımlık bir adım boyutu kullanarak 9 örtüşen zaman adımına bölündü. En sonunda, tek bir klibin nihai ses öznitelikleri, 9x40x40 şeklinde veri elde edildi. Her bir klibin süresi 5 saniye olduğundan, zaman adım sayısı da 9 olarak sabittir.

Tablo I. Modellere Göre Sonuçlar

| Sekans Modeller | | Sonuçlar | | | |
|---|---|---|---|---|---|
| | | Kesinlik | Duyarlılık | F1 Skor | Destek |
| Transformer | Kızgın | 0.727 | 0.743 | 0.735 | 140 |
| | Mutlu | 0.946 | 0.879 | 0.911 | 140 |
| | Nötr | 0.506 | 0.683 | 0.582 | 120 |
| | Üzgün | 0.497 | 0.614 | 0.550 | 140 |
| | İğrenmiş | 0.717 | 0.779 | 0.747 | 140 |
| | Korkmuş | 0.733 | 0.314 | 0.440 | 140 |
| | Harmonik Ortalama | 0.653 | 0.597 | 0.625 | - |
| LSTM | Kızgın | 0.720 | 0.807 | 0.761 | 140 |
| | Mutlu | 0.966 | 0.821 | 0.888 | 140 |
| | Nötr | 0.630 | 0.725 | 0.674 | 120 |
| | Üzgün | 0.539 | 0.736 | 0.622 | 140 |
| | İğrenmiş | 0.655 | 0.800 | 0.720 | 140 |
| | Korkmuş | 0.841 | 0.264 | 0.402 | 140 |
| | Harmonik Ortalama | **0.699** | 0.586 | 0.638 | - |
| GRU | Kızgın | 0.659 | 0.829 | 0.734 | 140 |
| | Mutlu | 0.897 | 0.936 | 0.916 | 140 |
| | Nötr | 0.634 | 0.650 | 0.642 | 120 |
| | Üzgün | 0.545 | 0.564 | 0.554 | 140 |
| | İğrenmiş | 0.655 | 0.814 | 0.726 | 140 |
| | Korkmuş | 0.804 | 0.321 | 0.459 | 140 |
| | Harmonik Ortalama | 0.680 | 0.604 | **0.640** | - |
| Maksiumum Havuzlama | Kızgın | 0.677 | 0.807 | 0.736 | 140 |
| | Mutlu | 0.909 | 0.714 | 0.800 | 140 |
| | Nötr | 0.533 | 0.675 | 0.596 | 120 |
| | Üzgün | 0.541 | 0.471 | 0.504 | 140 |
| | İğrenmiş | 0.593 | 0.686 | 0.636 | 140 |
| | Korkmuş | 0.664 | 0.507 | 0.575 | 140 |

| Sekans Modeller | Sonuçlar | | | |
|---|---|---|---|---|
| | Kesinlik | Duyarlılık | F1 Skor | Destek |
| Harmonik Ortalama | 0.632 | **0.620** | 0.626 | - |

*3. Deneysel Sonuç*

Deneylerimizde, bir grup büyüklüğü (batch size) olarak 16 kullanıyoruz ve Adam optimizer, 1e-4 başlangıç öğrenme hızıyla (initial learning rate) kullanılmıştır. Ağlar erken durdurma ile birlikte 15 devir (epoch) için eğitilmiştir. Kayıp fonksiyonu (Loss Function) olarak çapraz entropi (Cross Entropy) kullanıldı. Uygulamalar, Tensorflow kullanılarak Python'da yapılmıştır.

Veri kümesi, oyunculara bağlı olarak bölünmüştür. Bu tür bir bölünme test kümesinde daha genelleşmiş bir sonuç verir. Özetle yapay sinir ağ modeli besleyen oyuncunun verisi test ya da validasyon kümesinde kullanılmamıştır. Veri kümesi eğitim, validasyon ve test olarak bölünmüştür. Test ve validasyon setleri, her biri 10 konuşmacı verisine sahiptir. Geri kalan veriler, eğitim seti olarak kullanılmıştır. Sonuçlar kesinlik, duyarlılık ve F1 skor metrikleri kullanılarak raporlanmıştır.

## IV. SONUÇ

Bu çalışmada, Transformer, LSTM, GRU ve maksimum havuzlama içeren algoritmaların duygu tanımadaki etkinliğini araştırdık. Deneysel çalışmalarda, çok kipli (ses ve video) duygu tanımanın doğruluğu tartışılmış, derin sinir ağı modelleme araçlarıyla elde edilen performanslar karşılaştırılmış ve analiz edilmiştir. Karşılaştırma sonucunda üç model (Transformer, LSTM ve GRU) yaklaşık olarak birbirlerine yakın performans gösterirken, maksimum havuzlama yönteminin herhangi bir sekans modeli içermemesine rağmen diğer modellere benzer performans göstermesi dikkat çekicidir. Bu CNN çıktılarının zaman boyunca farklı modellerle beslenilse bile yerel minimumdaki değişiminin küçük olduğu gözlenmiştir. Ses ve görüntüler önce çok katmanlı CNN modelleri tarafından işlenmiştir ve hemen ardından bu modellerin çıktıları çeşitli sekans modellerine beslenmiştir. Sekans modeli olarak GRU, Transformer, LSTM ve Maksimum Havuzlama kullanılmıştır. Tablo I. ve Şekil 2. 'deki sonuçlara göre CREMA-D veri kümesinin karşılaştırma sonucunda, F1 skorda en iyi sonucu gösteren 0.640 ile GRU bazlı mimari, kesinlik metriğinde 0.699 ile LSTM bazlı mimari, duyarlılık da ise zaman içinde maksimum havuzlama bazlı mimari 0.620 ile en iyi sonucu göstermiştir. Sonuç olarak, karşılaştırılan modellerin birbirine yakın performans verdiği gözlenmiştir. Duygu tanımada kullanılan modellerin etkinliği kanıtlanmıştır.

Gelecekte yeni model yapıları araştırılıp farklı bakış açılarıyla duygu tanıma problemi yeniden ele alınılabilir. Sekans modellerinin yanı sıra öznitelik çıkarımı için farklı yaklaşımlar benimsenebilir. Ayrıca farklı kiplerin model çıktısına olan etkisi gözlemlenebilir.